# Pit-Pattern Classification of Colorectal Cancer Polyps Using a Hyper Sensitive Vision-Based Tactile Sensor and Dilated Residual Networks

Nethra Venkatayogi*, Qin Hu*, *Student, IEEE*, Ozdemir Can Kara*, *Student, IEEE*, Tarunraj G. Mohanraj, and S. Farokh Atashzar, *Member, IEEE*, Farshid Alambeigi, *Member, IEEE*

*Abstract*— In this study, with the goal of reducingthe early detection miss rate of colorectal cancer (CRC) polyps, we propose utilizing a novel hyper-sensitive vision-based tactile sensor called HySenSe and a complementary and novel machine learning (ML) architecture that explores the potentials of utilizing dilated convolutions, the beneficial features of the ResNet architecture, and the transfer learning concept applied on a small dataset with the scale of hundreds of images. The proposed tactile sensor provides high-resolution 3D textural images of CRC polyps that will be used for their accurate classification via the proposed dilated residual network. To collect realistic surface patterns of CRC polyps for training the ML models and evaluating their performance, we first designed and additively manufactured 160 unique realistic polyp phantoms consisting of 4 different hardness. Next, the proposed architecture was compared with the state-of-the-art ML models (e.g., *AlexNet* and *DenseNet*) and proved to be superior in terms of performance and complexity.

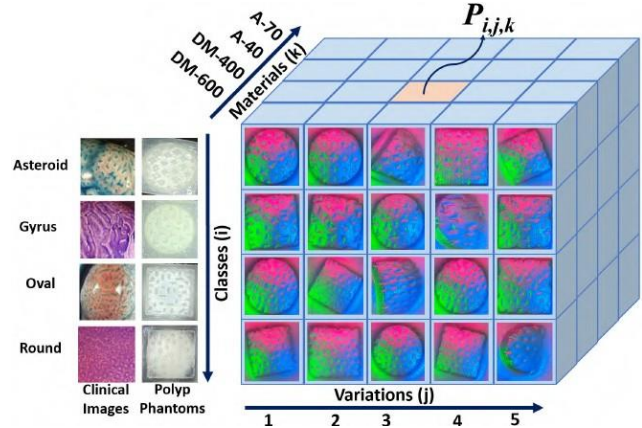

Fig. 1. A conceptual three-dimensional tensor showing the dimensions of the used image dataset collected by the HySenSe on the 5 out of 10 variations of 3D printed polyp phantoms.

## I. INTRODUCTION

Colorectal cancer (CRC) is the second most commonly diagnosed cancer and the leading cause of cancer-related deaths worldwide (i.e., over 935,000) in 2020 [1]. Early diagnosis of pre-cancerous (i.e., polyps) and cancerous lesions is, therefore, crucial and can increase the survival rate of CRC patients to 90% [2], [3]. The key step during the CRC screening process using a colonoscope involves the identification and differentiation of neoplastic lesions (i.e., lesions with high-cancerous potential) from non-neoplastic polyps (i.e., lesions with a low risk of developing cancer) that directly determines the management and treatment procedure of these lesions [2]. To perform this classification during the screening process, lesions' morphological characteristics (i.e., shape, size, and stiffness) and visual appearances are often considered and analyzed using a set of guidelines first introduced by Kudo et al. [4]. For instance, utilizing a magnified colonoscope, which allows for the visualization of intricate patterns on the mucosal surface and the vascular architecture, an approach called "*pit-pattern*" analysis is often used by clinicians for CRC screening. In this approach, the surfaces of lesions are inspected to identify various types of pit patterns, shown in Fig. 1, including the Type *I* (roundish), Type *II* (asteroid), Type *III* (tubular), and Type *IV* (dendritic/gyrus brain-like) [5]. Of note, among these patterns, *Type I* and *Type II* represent the normal/hyperplastic polyps and *Type III* and *Type IV* constitute the neoplastic variants [5]. As shown in Fig. 1, across patients, polyps often have a high degree of variation in morphological characteristics and visual appearances [6]. This often has led to polyp classification being evaluator-dependent as well as more difficult for clinicians to be able to differentiate between neoplastic and non-neoplastic polyps [7], [8]. Subsequently, these limitations have contributed to polyp miss rates as high as 20%, demanding effective diagnostic techniques to reduce the risks of mortality and improve chances of favorable prognosis [8].

To assist clinicians in the detection and classification of polyps and pit-patterns, recent innovations in colonoscopy have been focused on the development of intelligent clinical decision support systems and computer-aided diagnostics (CAD) [9], [10], [11], [12]. Primarily, the development of CAD systems for polyp detection and classification originated with computer vision methods such as hand-crafted feature extraction, wavelet-based methods, and creating energy maps [10], [12]. However, these methods have difficulties in characterizing patterns properly due to changes in illumination and the appearance of polyps in the images provided during colonoscopy [9]. When dealing with clas-

* These authors contributed equally to this work.

Research reported in this publication was supported by the University of Texas at Austin and MD Anderson Cancer Center Pilot Seed Grant.

Nethra Venkatayogi is the with Department of Biomedical Engineering at The University of Texas at Austin, Austin, TX, USA, 78712. email: (venkatayoginethra@utexas.edu)

Qin Hu, and S. Farokh Atashzar are with Department of Electrical and Computer Engineering, and Department of Mechanical and Aerospace Engineering, New York University, NY, USA. email: (qh503@nyu.edu, f.atashzar@nyu.edu)

Ozdemir Can Kara, Tarunraj G. Mohanraj and Farshid Alambeigi are with Walker Department of Mechanical Engineering and the Texas Robotics at The University of Texas at Austin, Austin, TX, USA, 78712. email: (ozdemirckara@utexas.edu, tarunrajgm@utexas.edu, and farshid.alambeigi@austin.utexas.edu)

sification based on the Kudo technique, deep learning using convolutional neural networks (CNNs) has been shown to be a promising solution. CNNs can act as global feature extractors that represent the total collection of intricate pit-pattern structures and, therefore, have great potential to increase the adenoma detection rate [9], [12]. Particularly, using the visual feedback provided by an endoscope, algorithms such as support vector machine (SVM), k-nearest neighbors (k-NN), ensemble methods, and convolutional neural networks (CNN) have been used for polyp detection and classification [11], [13], [14]. A detailed review of the ML-based CAD in a colonoscopy screening can be found in [11].

Aside from the mentioned algorithms, more advanced and complex architectures have also been utilized for CRC polyp classification including the Residual Networks (ResNet), Densely connected CNNs (DenseNet), and AlexNet [15], [12], [9], [16], [17], [18]. Among these, ResNet has shown to have notable generalization qualities due to its ability to ease the training of deep neural networks and reduce the effects of vanishing/exploding gradients [16], [19]. Nevertheless, deep neural networks, such as ResNet, tend to perform poorly on small datasets due to the occurrence of overfitting, thus impairing the model's translatability to clinical settings [20], [21]. Additionally, deep networks require a large, annotated, and balanced dataset, which can be challenging to obtain in the medical field [9].

To overcome the aforementioned barriers, and particularly the access to a rich and high-quality image dataset of CRC polyps' (with different types, shapes, textures, and sizes, which may also vary among cancer patients), *transfer learning* techniques have been proposed in the literature [9], [22]. This technique features the transfer of general image recognition subtasks (e.g., pattern identification and feature extraction) from large models trained on ImageNet to another, possibly smaller dataset even if differences are present in the source and target domains [9], [22]. Transfer learning has been utilized in many medical CAD systems, especially in polyp classification tasks [9], [23], [22]. Nevertheless, it should be noted that these works utilize datasets of over several thousand images and yet consider them as "*small*" datasets [20], [21]. Nevertheless, when utilizing novel technologies different from the typical colonoscopic images, access to datasets with the size of several thousand images or obtaining such images is extremely difficult and time-consuming [24]. To address this, *dilated convolutions* have been shown to help increasing the resolution of output feature maps while also improving the receptive field [25], [26]. This critical feature aids in capturing the intricate details that are crucial in pit-pattern classifications, where there can be occurrences of similar-looking images [21]. Moreover, dilated convolutions paired with transfer learning approaches have shown to have promising results when access to large datasets is limited [21], [20].

Aside from data access concerns, camera occlusions are another major drawback of existing colonoscopy screening. This critical limitation can result in low-quality or complete loss of video footage, failure of the ML-based CAD, and therefore overlook polyps at colon flexures during screening [27], [28]. To address this limitation, novel screening technologies are necessary that are insusceptible to camera occlusions.

To collectively address the aforementioned limitations of existing colonoscopy screening and ML-based CAD approaches and in order to reduce the early detection miss rate of CRC polyps using the pit-pattern classification, in this study, we propose utilizing (i) a novel hyper-sensitive vision-based tactile sensor (called HySenSe) and (ii) a complementary and novel ML-based CAD architecture. The proposed architecture explores the potentials of utilizing dilated convolutions, the beneficial features of the ResNet architecture, and the transfer learning concept applied on a small dataset with the scale of hundreds of images. The developed tactile sensor provides high-resolution 3D textural images of CRC polyps that will be used for their accurate classification via the proposed dilated residual network. To collect realistic surface patterns of CRC polyps for training the ML models and evaluating their performance, we first designed and additively manufactured 160 unique realistic polyp phantoms consisting of 4 different hardness. Next, the performance of the proposed architecture compared with the other common ML approaches was quantitatively evaluated and compared using various clinically relevant statistical metrics.

## II. METHODOLOGY

### A. Hyper Sensitive Vision-Based Tactile Sensor

To accurately classify the pit-pattern of CRC polyps based on the high-quality image dataset, we have developed a Hyper Sensitive Vision-based Tactile Sensor called HySenSe [29]. As shown in Fig. 3 and described in [29], HySenSe provides high-resolution 3D textural images in very low interaction forces (i.e., < 2$N$), without sacrificing its durability and fidelity. Of note, the interaction force between CRC polyps and HySenSe is significantly lower than the evidenced force threshold [30], which ensures the safety of the procedure.

As illustrated in Fig. 3 (8), HySenSe consists of a dome-shape elastic silicone layer that directly interacts with an object (i.e., a deformable polyp), a transparent acrylic layer that provides support to the gel layer, and a camera that faces upwards to observe the movements of the gel layer and is fixed to the 3D printed rigid frame, and an array of Red, Green, and Blue LEDs placed 120 degrees apart for illumination, aiding in the recreation of the 3D textural features when an object interacts with the sensor. To manufacture HySenSe, we used a 5 MP camera (Arducam 1/4 inch 5 MP sensor mini) that was fixed to a 3D printed rigid frame fabricated using a high-resolution printing technology (Formlabs Form 3, Formlabs Inc.) and the clear resin material (FLGPCL04, Formlabs Inc.). Based on the camera focal length and the field of view, the height of the rigid frame was selected as $h_s$=24 mm. Also, to sufficiently illuminate the sensor and create contrast for the output images of the HySenSe, we utilized an array of Red, Green, and Blue LEDs (WL-SMTD Mono-Color 150141RS63130, 150141GS63130, 150141BS63130, respectively) placed 120 degrees apart.

For the fabrication of the HySenSe gel layer, as the critical component of the sensor interacting with the polyps, we used a soft transparent platinum cure two-part (Part A- Part B) silicone (P-565, Silicones Inc.). The mixture mass ratio of the silicone was selected as 14:10:4 (A:B:C), in which Part C represents phenyl trimethicone softener (LC1550, Lotioncrafter). Before pouring the silicone mixture into the silicone mold (chocolate mold, Baker Depot) with R= 35 mm, we coated the surface of the mold with Ease 300 (Mann Technologies) in order to prevent adhesion. After 10-15 minutes, mixtures were poured into a silicone mold and then placed into a vacuum chamber to complete the degassing process for removing the trapped bubbles in the mixture. Later, the degassed silicone mixture was cured and solidified in a curing station (Formlabs Form Cure Curing Chamber).

In order to increase the sensitivity and durability of the sensor, compared with similar vision-based tactile sensors (e.g., GelSight [31]), we modified the fabrication procedure of the sensor by utilizing Specialty Mirror Effect spray paint (Rust-Oleum Inc.) as the coating of the fabricated gel layer Particularly, in our updated procedure, instead of brushing the surface of the gel layer with an aluminum powder, we sprayed the surface 5 times with 1-minute intervals and fabricated a dramatically higher sensitive sensor. Of note, the increased sensitivity of HySenSe mitigates the need for applying a high interaction force to the polyp phantoms in order to obtain high-resolution textural images. This important feature not only reduces the typical fabrication steps of a vision-based tactile sensor, but it also prevents damaging the sensor and ensures the safety of the procedure when measuring the texture of the CRC polyps. Finally, as the last fabrication step, instead of using a thin layer of silicon mixture with grey pigment to prevent light leakage (adding another intermediate fabrication step), a thin layer of silicone mixture, with the same mixture ratio, was poured on the surface of the gel layer to cover the spray coating. Of note, the hardness of the fabricated gel layer was measured as 00-18 using the Shore 00 scale durometer (Model 1600 Dial Shore 00, Rex Gauge Company). Fig. 2 clearly illustrates the exceptional sensitivity of the HySenSe in comparison to the GelSight sensor experimented under identical conditions.

### B. Polyp Phantoms

As illustrated in the three-dimensional tensor shown in Fig. 1, the utilized dataset can be visualized as composed of polyp phantoms that were designed and additively manufactured by varying the indices $(i, j, k)$ along each dimension of the tensor. Each unique polyp $P_{i, j, k}$ is characterized by its Kudo class $i$ (called Asteroid (A), Gyrus (G), Oval (O), and Round (R) throughout the paper, respectively), the geometric textural variation $j$, and the material $k$, which represents varying stiffness characteristics [2], [32]. The textural proper-ties of each polyp class and its corresponding Kudo classification, inspired by Kudo et al., are summarized in Fig. 4a [4]. Of note, all of the fabricated polyp types have an average spacing of 600 microns between neighboring features and a uniform pit depth of 500 microns. Additionally, based on [4], a deviation of around $\pm 100 - 150$ $\mu m$ was included in the major axis, minor axis, and spacing parameters to better simulate realistic texture qualities. After the design of polyp phantoms in SolidWorks software (SolidWorks, Dassault Systèmes), each of the $4 \times 10 \times 4 = 160$ polyps were printed using the J750 Digital Anatomy Printer (Stratasys, Ltd). The polyp phantoms were printed with different hard and soft materials. Table I summarizes the measured shore hardness of these phantoms obtained with an electronic Shore A scale durometer (Starret 3805B).

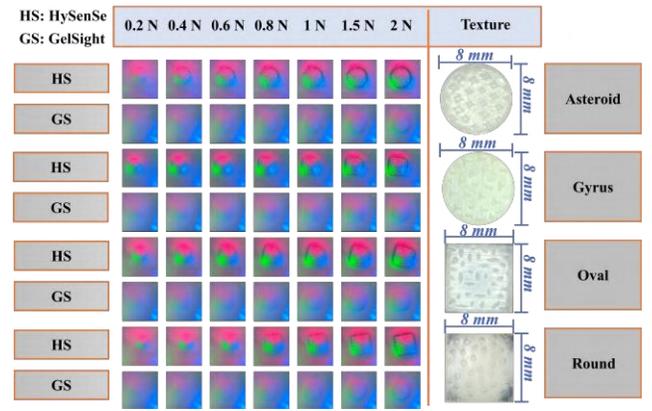

Fig. 2. Progression of the visual outputs for the HySenSe and GelSight sensors. Each two row represents the different types of CRC polyps used for experiments. Also, the top row indicates the applied forces corresponding to each image.

### C. Experimental Setup and Procedure

Fig 3 shows the experimental setup prepared to record the HySenSe's outputs while measuring the textural images of each unique polyp. The setup consists of HySenSe, a linear stage (M-UMR12.40, Newport) with 1 $\mu m$ precision that was used to attach and push polyps on the HySenSe's deformable gel layer, a digital force gauge (Mark-10 Series 5, Mark-10 Corporation) with 0.02 N resolution– used to measure the interaction forces between the gel layer and each polyp, and a Raspberry Pi 4 Model B– used for the video recording and data streaming and further image analysis. MESUR Lite data acquisition software (Mark-10 Corporation) was utilized to record the forces. To perform experiments and simulate realistic interactions between the sensor gel layer and the polyps, two orientations of the linear stage were utilized, angled at 0° and 45°. Of note, 0°mimics a complete interaction between the gel layer and the polyp in which the whole texture of polyp can be captured whereas the 45° simulates a situation in which part of the polyp's texture can be measured by the HySenSe. Each of the unique polyps was vertically pushed onto the HySenSe until a 2 N force was exerted in the 0° orientation. Additionally, 5 geometric variations ($j$) from each polyp class ($i$) across each of the four materials ($k$) were used for the experiments in a 45° orientation of the polyp relative to the surface of HySenSe, again up to a 2 N force, for a total of 80 angled experiments. 11 experiments were disregarded due to poor image quality during experimentation, resulting in a total of 229 experiments that were performed across the two orientations

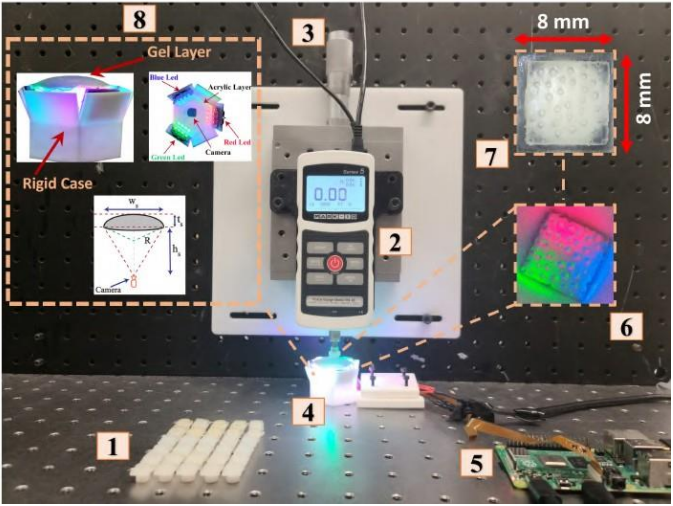

Fig. 3. Experimental Setup: 1- Fabricated polyp phantoms, 2- Mark-10 Series 5 Digital Force Gauge, 3- M-UMR12.40 Precision Linear Stage, 4- Our proposed novel HySenSe sensor, 5- Raspberry Pi 4 Model B, 6- HySenSe image output, 7- Dimensions of polyp phantom, 8- HySenSe top view, side views, and visualization of HySenSe dimensions, $h_s$ is the height of the 3D printed rigid frame, $t_s$ is the thickness of the gel layer $w_s$ is the width of the gel layer, and R is the radius of the dome-shaped gel layer.

TABLE I
MATERIAL HARDNESS PROPERTIES FOR POLYP PHANTOMS

| Material | Shore Hardness |
| --- | --- |
| DM400 (Agilus30 Clear + TissueMatrix) | A 1-2 |
| DM600 (Agilus30 Clear + TissueMatrix) | A 30-40 |
| A-40 (Agilus30 Clear + Vero Pure White) | A 40 |
| A-70 (Agilus30 Clear + Vero Pure White) | A 70 |

### D. Dataset and Pre-processing steps

Across the 229 experiments, the class counts for polyps A, G, O, and R were 57, 57, 55, and 60, respectively. The frame associated with the HySenSe output at 2 N was extracted and used for the creation of the dataset. From the 229 unique polyp images, cross-validation was performed with a 5-fold Stratified K-Fold for the 80% and 20% training and testing splits, and within the training split, 80% and 20% were further split into training and validation datasets, respectively, for each fold. In total, the training and test datasets consisted of 182 and 47 images, respectively. The HySenSe output was manually cropped to only include the polyp area of interest and resized from the 1080 × 1280 pixels to 224 × 224 pixels in order to have a constant input size for all models. To further improve the generalization ability of the algorithm, various geometrical transformations such as random cropping, vertical and horizontal flips, and random rotations between -45° to 45°were included as augmentations, each with an independent occurrence probability of 0.5. Examples of rotations that occurred through experiments and images after the applied transformations are visualized in Fig 4. Of note, input normalization was not used to preserve the original raw color mapping of the textural images provided by the HySenSe.

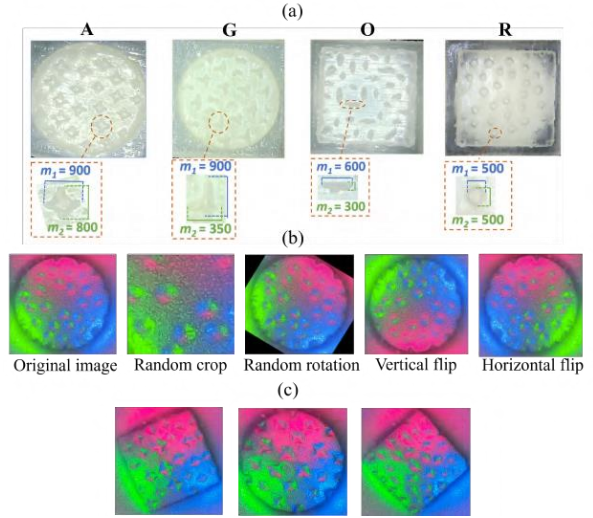

Fig. 4. Example images of HySenSe-outputs (a) Dimensions of fabricated polyps (b) Geometric transforms utilized (c) Rotational variations that resulted during experiments.

### E. Model selection

The ResNet model can be considered as the state-of-the-art architecture in polyp detection [27], [33]. The ResNet architecture reduces the effects of vanishing/exploding gradients by behaving like an ensemble of small weakly dependent networks [16], [19]. Additionally, ResNets have skip connections that help with faster convergence and contribute to solving the degradation problem, where a model's performance decreases with complexity and depth [16]. The standard ResNet convolves with non-dilated kernels. Dilated kernels keep the spatial resolution of feature maps with fewer trainable parameters and enhance the receptacle field of the network [34]. Additionally, Yu et al. [26] demonstrated the effectiveness of dilation, particularly in residual networks. Dilation has also been effectively utilized in understanding high-density images such as highly congested scenes [35]. Considering the high textural density of the features in our dataset and in colonoscopic images, dilation secures great potential when dealing with intricate features and has also been previously utilized in polyp classification tasks [21].

Thus, in our proposed architecture, we impose convolutional dilation to reduce the complexity of the proposed model and prevent overfitting, which is crucial to a small dataset. In this regard, the inclusion of dilation in the proposed model has the capability to (1) increase the receptive field of CNNs to capture multi-scale contextual information, which is critical to polyp data, (2) prevent overfitting when using deep networks, and (3) improve performance without increasing the model complexity.

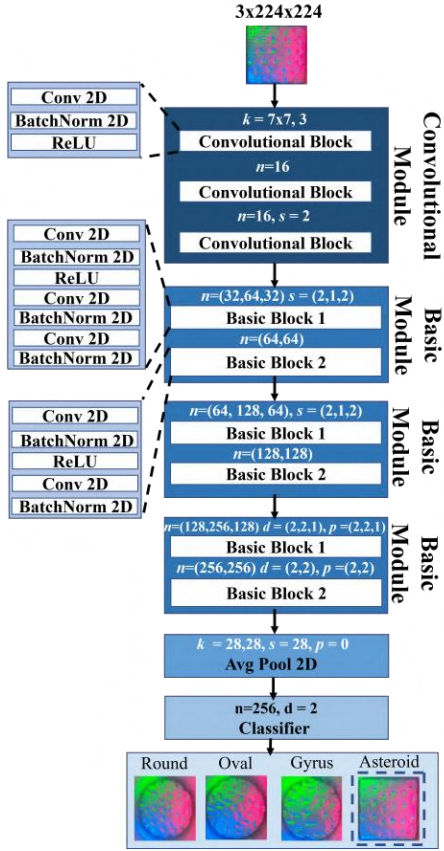

Fig. 5. Model structure. The numbers specify the number of input channels and layer-wise values for parameters of $n$, $k$ $d$, $s$, and $p$ for each block, denoting the number of filters, kernel sizes, dilation, strides, and paddings, respectively. Multiple values are shown in a block corresponding to each of the block's convolutional layers. Default values include $k$=3, $p$=1, $d$=1, $s$=1, unless stated in the figure otherwise. The dashed box indicates the predicted class.

In addition to the above-mentioned points, we remove the max-pooling layers from the standard ResNet. This is mainly due to the fact that a max-pooling layer summarizes the information of pooling regions in a feature map by preserving the pixel with the maximum value from each region. Although a max-pooling layer helps enhance the invariance to position changing and lighting conditions, it reduces the resolution of the feature map and can cause information lost on distinguishing features if most of the pixels within a pooling region have high magnitudes [36]. Therefore, we remove the max-pooling layers.

The proposed model for our purposes is illustrated in Fig. 5 and features (1) *a convolutional module* composed of three convolutional blocks, each composed of a 2D convolution layer, a Batch Normalization (BN) layer, and a Rectified Linear Unit (ReLU) layer, (2) *three Basic ResNet modules* composed of two types of Basic ResNet blocks, one having a convolutional block followed by two additional convolutional blocks without a ReLU activation, while the other has two convolutional blocks, with the second not utilizing a ReLU activation, (3) *an average pooling layer*, and (4) *a classifier* composed of a 3×3 convolutional layer that outputs to four classes. The last Basic ResNet block contains dilated convolutions and padding both with a factor of 2. Additionally, the use of padding along with dilations in this last block allows for better capturing of information on the edges. The proposed model takes advantage of transfer learning, initialized with pre-trained weights optimized based on ImageNet, followed by a fine-tuning process that further adjusts the model weights to complement our dataset.

To evaluate our model's performance relative to other commonly used architectures for polyp classification tasks, we utilize the first 3 blocks of ResNet-18 (named *LightResNet*) to be comparable to the proposed model, in addition to *DenseNet*, and *AlexNet* [16], [17], [18]. Additionally, to compare the computational complexity and size/depth of each model, the number of trainable parameters is monitored. A model with a greater number of parameters is typically more complex as a larger number of computations need to be executed to optimize the internal parameters for a given dataset, resulting in higher training times and costs [37].

*F. Evaluation metrics*

To compare the performance of the proposed model with the other CAD models, we used different evaluation metrics including the accuracy ($M_A$), sensitivity ($M_{Sens}$), specificity ($M_{Spec}$), precision ($M_{Prec}$), F1-score ($M_{F1}$), and AUC ($M_{AUC}$). These metrics are especially relevant to evaluate the performance of the model for the CRC polyp classification and provide more comparisons beyond accuracy. Primarily, accuracy represents the model's number of correct predictions across all classes. In addition, it is crucial to ensure that the model's true positive ($M_{Sens}$) and true negative ($M_{Spec}$) predictions are high to avoid the classification of a neoplastic lesion as a non-neoplastic lesion and, therefore, reduce the unnecessary need for performing a biopsy of a non-neoplastic polyp. Sensitivity (i.e. recall) was the most commonly used metric to evaluate the performance of deep learning models for CRC polyp classification [12]. Moreover, $M_{Prec}$ is a measure representing the confidence of the model in predicting a type of tumor. The F-1 score is also important to consider when there are imbalances in a model's sensitivity and precision values. Lastly, AUC is critical to consider as it is the probability the model will score a randomly chosen neoplastic polyp higher than a randomly chosen non-neoplastic polyp, and it is optimal to have a value close to 1. For further analysis, confusion matrices were also calculated and normalized by taking the average across the row, or computing class-wise recall values using the formula below

$$C_M{}^{i,j} = \frac{|y_i \cap \hat{y}_j|}{|y_j|} \tag{1}$$

where $y$ is the set of true labels, $\hat{y}$ is the set of predicted labels, and $i$ and $j$ refer to the predicted and true class labels.

*G. Hyperparameters*

The hyperparameters for each model and its respective optimizer, as shown in Table II, were chosen such that they maximize the performance metrics of the model on the aforementioned 20% test split. Our proposed model and LightResNet were each trained for 150 epochs, while DensetNet and AlexNet were both trained for 100 epochs. The batch size was maintained at a uniform value of 4 for all the models. This value was observed to produce the best performance across all the compared models.

TABLE II
HYPERPARAMETERS FOR THE UTILIZED MODELS

| Model | Epochs | Batch Size | $LR_1$ | $LR_2$ |
|---|---|---|---|---|
| Proposed Model | 150 | 4 | 0.001 | 0.01 |
| DenseNet | 100 | 4 | 0.0001 | 0.01 |
| AlexNet | 100 | 4 | 0.0001 | 0.01 |
| LightResNet | 150 | 4 | 0.001 | 0.01 |

TABLE III
EVALUATION METRICS OF THE UTILIZED MODELS

| Metrics | Proposed | DenseNet | AlexNet | LightResNet |
|---|---|---|---|---|
| Validation Acc. (%) | 97.33 | 96.89 | 82.55 | 94.61 |
| Test Acc. (%) | 93.62 | 95.84 | 77.06 | 89.79 |
| Sensitivity (%) | 97.33 | 95.64 | 84.85 | 91.10 |
| Precision (%) | 98.08 | 96.43 | 89.09 | 93.08 |
| Specificity (%) | 99.30 | 98.57 | 95.00 | 97.14 |
| F1-score (%) | 97.90 | 96.03 | 86.90 | 92.08 |
| AUC | .9976 | .9994 | .9494 | .9901 |
| Parameters (mil) | 2.8 | 7.2 | 62 | 2.8 |

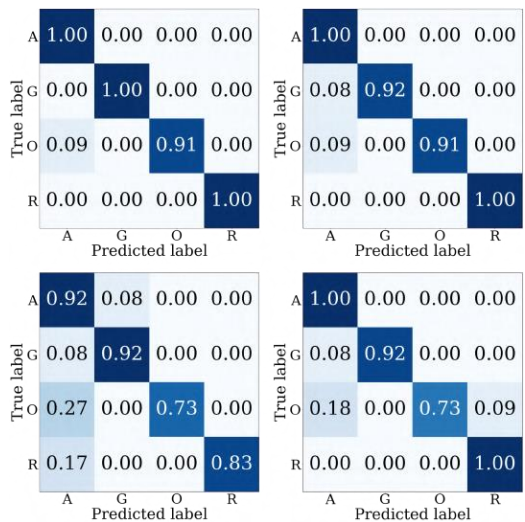

Fig. 6. Normalized confusion matrices of the utilized models (from left to right and top to bottom): The proposed model, *DenseNet*, *AlexNet*, *LightResNet*

All our models were optimized using the AdaBound optimizer. AdaBound is a variation of the conventional Adaptive Moment Estimation (Adam) method that utilizes dynamic learning rates consisting of a lower bound and an upper bound, which converge to a constant value through training [38]. The dynamic approach enables AdaBound to behave like the standard Adam optimizer at the beginning of training and later smoothly transform to Stochastic Gradient Descent (SGD), incorporating characteristics of both of these common optimization methods. $LR_1$ is the initial learning rate, while $LR_2$ is the learning rate at which AdaBound transforms to SGD.

III. RESULTS AND DISCUSSION

The evaluation metrics, as reported in Table III, were recorded and averaged across the 5-folds for all the models. The validation accuracy was computed using the 20% validation split within the 80% training split, while all the other metrics were computed using the initial 20% test split not used for the training process. Confusion matrices were normalized and reported in Fig. 6. As seen through the metrics in Table III, our proposed model outperforms *AlexNet*. *DenseNet*, and *AlexNet* can only achieve reasonable performance with 2.6× and 22 × more trainable parameters relative to the proposed model, respectively. Additionally, leveraging the benefits of convolutional dilation with padding, our proposed model achieves considerably better metrics and a smaller gap between training and test accuracy compared with the comparable *LightResNet*, exemplifying the proposed model's performance and generalization. Furthermore, during training, the *AlexNet* model was observed to overfit; the large scale of this model, and thus the large parameters-to-feature-complexity ratio, resulted in *AlexNet* being unfeasible for our application.

As seen in the confusion matrix for our proposed model, the sensitivity for the non-neoplastic lesions is at 100% while the specificity is of a comparatively high value of 99.30%, signifying the model's potential to correctly identify true negatives, therefore reducing the number of biopsies necessary to characterize the cancerous potential of the polyp. Additionally, it should be noted that our lowest performance occurred while predicting the neoplastic O class, but this is still at an acceptable 91%.

It is crucial to consider the trade-off between a model's performance and its size/depth to find the optimal model that produces the best metrics while also achieving computational efficiency. The performance of the proposed model indicates that the inclusion of dilated convolutions in just one basic ResNet module enables it to outperform the compared models, which do not utilize dilated convolutions in their respective architectures.

IV. CONCLUSION

In this paper, with the goal of reducing the early detection miss rate of CRC polyps, we presented an AI-driven approach for a novel hyper sensitive vision-based tactile sensor, called HySenSe. The proposed deep learning model decoded the textural information provided by the HySenSe to reliably differentiate between various types of pit-patterns of neo-plastic and non-neoplastic CRC polyps. The core of the proposed CAD is a dilated CNN that was designed and trained through a transfer learning approach and custom-designed realistic CRC polyp phantoms. The proposed model combined the benefits of dilated convolutions and transfer learning when using a small dataset of 229 images captured in this study by the HySenSe system. The model was compared with state-of-the-art models, such as *AlexNet*, and proved to be superior in terms of performance and complexity.